\PassOptionsToPackage{numbers,compress}{natbib}
\documentclass{article}

\usepackage[sglblindworkshop, final]{ai4nextg_neurips_2025}

\pdfoutput=1

\usepackage{amsmath}
\usepackage[pdftex]{graphicx}
\usepackage[utf8]{inputenc} 
\usepackage[T1]{fontenc}    
\usepackage{booktabs}       
\usepackage{amsfonts}       
\usepackage{nicefrac}       
\usepackage{microtype}      
\usepackage{xcolor}         
\usepackage{amsmath}
\usepackage[utf8]{inputenc} 
\usepackage[T1]{fontenc}    
\usepackage{subcaption}
\usepackage[hyphens,spaces,obeyspaces]{url}
\usepackage{graphicx} 
\usepackage{siunitx}
\usepackage{hyperref}

\newcolumntype{x}[1]{>{\centering\arraybackslash\hspace{0pt}}p{#1}}

\title{Through the telecom lens: Are all training samples
important?  }

\author{%
\textbf{Shruti Bothe}$^{*}$, \textbf{Illyyne Saffar}$^{\dagger}$, \textbf{Aurelie Boisbunon}$^{\dagger}$, \textbf{Hasan Farooq}$^{*}$,\\
\textbf{Julien Forgeat}$^{*}$, \textbf{Md Moin Uddin Chowdhury}$^{*}$\\[0.5em]
Ericsson Research\\
Santa Clara, USA$^{*}$ \quad Massy, France$^{\dagger}$\\[0.5em]
\begin{tabular}{c}
\texttt{shruti.bothe, illyyne.saffar, aurelie.boisbunon, hasan.farooq,}\\
\texttt{julien.forgeat, md.moin.uddin.chowdhury@ericsson.com}
\end{tabular}
}

\begin{document}

\maketitle

\begin{abstract}
  The rise of AI in telecommunications, from optimizing Radio Access Networks to managing user experience, has sharply increased data volumes and training demands. Telecom data is often noisy, high-dimensional, costly to store, process, and label. Despite AI’s critical role, standard workflows still assume all training samples contribute equally. On the other hand, next-generation systems require AI models that are accurate, efficient, and sustainable. This paper questions the assumption of equal importance by focusing on applying and analyzing the roles of individual samples in telecom training and assessing whether the proposed method optimizes computation and energy use. We perform sample-level gradient analysis across epochs to identify patterns of influence and redundancy in model learning. Based on this, we propose a sample importance framework that selectively prioritizes impactful data and reduces computation without compromising accuracy. Experiments on three real-world telecom datasets show that our method preserves performance while reducing data needs and computational overhead while advancing the goals of sustainable AI in telecommunications.

\end{abstract}

\section{Introduction}
From one generation to the next, leading up to 6G, the telecommunications industry is undergoing a major shift driven by the convergence of AI and next-generation network design, with ML models increasingly deployed for critical functions such as user traffic prediction, beamforming optimization, anomaly detection, and intelligent handover control \cite{gote20255g}. Although these models improve automation and efficiency, their deployment brings significant challenges, including high demands for data, compute, and training time.

The telecommunications industry is increasingly focused on energy efficiency as operational costs and environmental impact become critical considerations \cite{10872842}. Dense 5G and future 6G networks consume substantial energy across base stations and core processing \cite{GSMA}, making reductions essential for lowering expenses and carbon footprint \cite{10872846}. Most efforts target specific use cases (e.g., base-station optimization or transmission protocols), while far fewer address the dual challenge of optimizing both AI training and application-level energy. This paper addresses that gap with an integrated strategy that considers AI model efficiency alongside network operational energy.

Meanwhile, the AI community has also recognized the environmental cost of large-scale training. Initiatives like \textit{Green AI} \cite{schwartz2020green} call for models that are not only accurate, but also compute- and energy-efficient. In telecommunications, this challenge is compounded by the sheer scale of networks, the high cost of operations, the dynamic nature of traffic, and the need to retrain frequently. Thus, a key question emerges:

\textit{"Are all training samples equally valuable in the context of telecom model training, or can we train smarter by focusing on the most impactful data?"}

A notable assumption in ML pipelines is that all training samples are equally important for model convergence and generalization. This assumption originates from traditional domains such as image classification or Natural Language Processing (NLP), where data distributions are often static and well-curated \cite{bottou-bousquet-2008}. In contrast, telecom data is fundamentally different: it is non-stationary, domain-specific, and frequently reflects rare but critical events. Moreover, many ML pipelines in telecom must operate under real-time or near-real-time constraints, making data curation, training efficiency, and sustainability major practical concerns.

This paper takes a gradient-centric view of sample utility in  model training with a focus on telecom applications. By analyzing how the gradients of individual samples evolve over training epochs, we categorize samples into those that are critical, redundant, or even detrimental to performance. Our central insight, through experimentation is that a significant portion of the training set offers marginal returns in terms of model improvement that suggests an opportunity for selective training that is both faster and more resource-efficient. To implement this notion, we introduce a sample importance framework. This model dynamically scores and filters samples during training based on their gradient norms, which serve as a proxy for influence on the loss incurred. In standard ML pipelines, the full dataset is used indiscriminately through randomized mini-batching, shuffling without regard to temporal structure, sequentially iterating over all samples in fixed-size batches, oversampling already well-represented periods, or repeatedly training on stable low-variance intervals. This approach focuses instead on selectively training and prioritizing samples that drive meaningful parameter updates. We validate this approach on multiple telecom datasets and show that models trained using only a fraction of the data can achieve comparable performance to those trained on the full dataset while reducing training time and compute cost.

The key contributions of this paper are: {(i) an analysis of} per-sample gradient dynamics across training epochs using a real-world open telecom dataset, {showing} that depending on the dataset size, the influence of a significant portion of training data contributes minimally to loss reduction or generalization; {(ii) the proposition of} a lightweight, model-agnostic mechanism to estimate sample importance during training based on gradient norm behavior that dynamically filters or re-weights data based on gradient importance. Unlike influence functions or forgetting metrics, our approach does not require retraining or access to ground truth labels; {and (iii)} Our framework supports goals of energy-efficient AI development by reducing redundant computation and data usage during training, {which} is especially relevant for large-scale network operators seeking to deploy AI models across distributed infrastructure.

\section{Related Works}

Research in core-set selection \cite{sener2018active} and influence functions \cite{koh2017understanding}  have explored ways to identify the most impactful samples, but these methods often come with high computational costs. Influence functions, for instance, measure how changing a sample’s weight affects model performance without retraining, yet their applicability to deep models is limited by convexity assumptions and the expense of computing Hessian inverses \cite{basu2020influence}. \cite{toneva2018an} proposed tracking “forgetting events” to reveal persistently misclassified or unstable samples, offering insights into data difficulty or noise. However, most of this work focuses on vision and NLP benchmarks, with little attention to structured, operationally constrained telecom data.

Other approaches, like curriculum learning \cite{bengio2009curriculum}, suggest ordering training samples by difficulty to speed learning. While effective in some domains, defining “difficulty” in telecom is challenging due to noisy or missing labels. Bothe et all \cite{curri_telecom} used curriculum learning to order training samples, but their work uses the full dataset for training. Meanwhile, AI in telecommunications has expanded rapidly, powering applications from network traffic forecasting \cite{Chen2021TrafficForecasting}  to load balancing, radio access network (RAN) tuning, and fault prediction. Yet, model training is often treated as a black box, with little examination of how training data characteristics affect reliability, retraining frequency, or energy use. As networks shift toward dynamic architectures like Open-RAN and edge-cloud splits, models must adapt to continuously evolving data streams by making efficient retraining not just beneficial, but necessary. Environmental costs are also rising. Recent studies \cite{strubell2020energy,schwartz2020green,luccioni2024power} warn of the carbon footprint of large-scale deep learning, but domain-specific efficiency strategies for telecom remain rare. This is especially important given that telecom data is costly to collect, privacy-sensitive, and often subject to real-time constraints.

Our work addresses these challenges with a lightweight, gradient-based sample importance metric that directly ties to operational KPIs. By allowing the model’s own learning dynamics to prioritize training data, we reduce computational overhead without compromising accuracy. This aligns AI efficiency with telecom’s operational, financial, and sustainability goals by making retraining faster, greener, and better suited to the demands of large-scale telecom networks.

\section{Problem Statement and proposed approach}

{Let $X\in\mathbb{R}^{t \times d}$ be a time series of length $t$ with $d$ features, and $Y$ be the target to predict. In the case of forecasting, $Y\in\mathbb{R}^{t' \times d}$ represents the same time series as $X$ with a given delay, which can have the same length $t$ or a different one. In the case of a classification problem, $Y\in\{1, \dots, c\}$ represents the class of the time series, such as the type of activity of a signal (SMS, call, internet), where $c$ is the number of classes.} The dataset $(x_i, y_i)_{i=1}^n$ denotes the collection of observations of $(X, Y)$ of size $n$.

{Let $f$ be the prediction model mapping $X$ to $Y$, and $\mathcal{L}:(\mathcal{X}\times\mathcal{Y})\mapsto \mathbb{R}$ be a loss function based on the predictions errors (e.g., the mean squared error in regression, or the cross-entropy in classification). We focus in this study on deep neural network architectures $f_\theta$, where $\theta\in \Theta$ denotes the weights of the neural network that are optimized through $\mathcal{L}$: $f_{\theta^*}=\arg\min_{\theta\in\Theta}\mathcal{L}(X,Y)$.

\subsection{Gradient Norm Computation for Importance score}
To identify important samples, we track per-sample gradient norms across all epochs. At epoch $e$ and for sample $s$, the gradient norm is computed as:
\begin{equation}
g_{e,s} = \sqrt{ \sum_{j=1}^P \left\| \frac{\partial \mathcal{L}_{e,s}}{\partial \theta_j} \right\|_2^2 },
\end{equation}
where $\theta_j$ denotes the $j^{\text{th}}$ trainable parameter in the network and $P$ is the total number of parameters.
These norms are stored in a matrix $\mathbf{G} \in \mathbb{R}^{E \times N}$, where $E$ is the number of epochs. The \textit{importance score} of a sample $s$ is then defined as:
\begin{equation}
{\mathcal{I}(s)}= \frac{1}{E} \sum_{e=1}^E g_{e,s}.
\label{eq:importance-score}
\end{equation}

\subsection{Important Sample Selection} 

{The goal of sample selection is to find a metric that orders the samples according to their importance, and to use that metric to select the most relevant ones for training the model. } In this work, the metric we consider is the importance score based on the gradient, defined in \eqref{eq:importance-score}.

The selection of the most $p$\% important samples thus corresponds to the set:
\begin{equation}
S^*_p = \underset{S \subset D ; |S| \leq k}{\arg\max}  \sum_{s\in S}\mathcal{I}(s),
\label{eq:selected-set}
\end{equation}
where $|S|$ denotes the cardinality of $S$ and $k = \left\lceil \frac{p}{100} \times n \right\rceil$ is the number of samples corresponding to $p$\%.

\section{Experiments and Results}

\subsection{Datasets and pre-processing}
\begin{enumerate}
    \item \textbf{Telecom Italia Big Data Challenge}: We use the publicly available \textit{Telecom Italia} dataset \cite{barlacchi2015multi}, which provides anonymized measurements of SMS, call and Internet activity in the city of Milan. For this work, we select the "Internet\_Activity" feature as both the input and output in a univariate forecasting task. An example of the shape of the univariate time series is shown in Figure~\ref{fig:pattern_a}. Data is normalized in the range [0, 1] to stabilize neural network training and avoid scale-induced gradient instability. The dataset is chronologically split 80-20 into training and testing subsets. No shuffling is applied to preserve temporal order and prevent look-ahead bias. The supervised learning representation is constructed using a lag-1 sliding window:
    \begin{equation}
    X_t = [x_t], \quad y_t = x_{t+1}
    \end{equation}
    where $x_t$ denotes the Internet activity at time $t$. The resulting samples are reshaped to $(n_{\text{samples}}, n_{\text{timesteps}}, n_{\text{features}})$ format, where $n_{\text{timesteps}} = 1$ and $n_{\text{features}} = 1$.
    
    \item \textbf{Proprietary data from a Telecom Vendor}: The second dataset consists of multivariate time series data collected every 15 minutes from 249 base station sites, covering both LTE and 5G New Radio (NR) technologies. It includes performance counters related to key KPIs such as physical resource block utilization, data volume, number of connected users, active sessions, throughput, and signaling overhead, with measurements from both uplink and downlink. Performance counters at the cell level are aggregated to match energy consumption data measured at the physical hardware level, requiring data pre-processing and mapping between cells and radio units. When multiple cells share a radio unit, their performance counters are aggregated accordingly and combined with corresponding energy consumption values, as shown in Figure~\ref{fig:pattern_b}. The dataset focuses on a subset of sites selected from a European city center and 50 neighboring sites within roughly five kilometers. 
    
    \item \textbf{5G Beam Selection data}: The 5G Beam Selection dataset \cite{Klautau18} is a synthetic 5G mmWave MIMO dataset generated by combining Simulation of Urban Mobility (SUMO) traffic simulator \cite{SUMO2018} with the Wireless InSite\textregistered~ray-tracing tool to capture realistic vehicular mobility and wireless propagation in an urban canyon in Rosslyn, Virginia, where a roadside unit transmits to 10 moving vehicles, producing 116 episodes with 50 time-sampled scenes each. For every scene, detailed ray-level channel information is recorded, including complex gains, power, angles of arrival/departure, time of arrival, and interaction types (line-of-sight, reflections, scattering), along with a 2D occupancy grid that encodes vehicle and receiver positions. 
\end{enumerate}

\begin{figure}[htbp]
    \centering
    \begin{subfigure}{0.48\linewidth}
        \includegraphics[width=\linewidth]{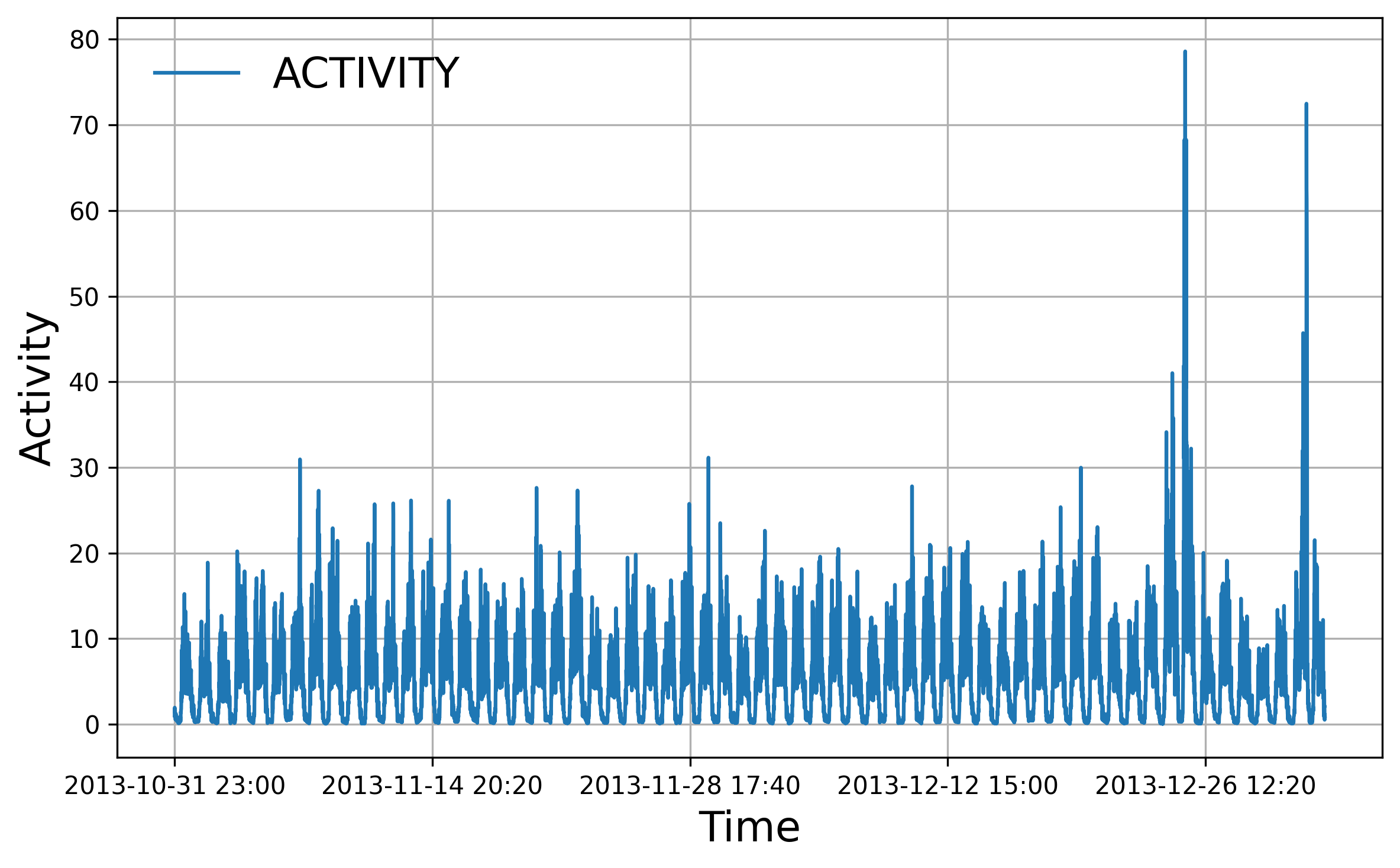}
        \caption{Telecom Italia}
        \label{fig:pattern_a}
    \end{subfigure}
    \hfill
    \begin{subfigure}{0.48\linewidth}
        \includegraphics[width=\linewidth]{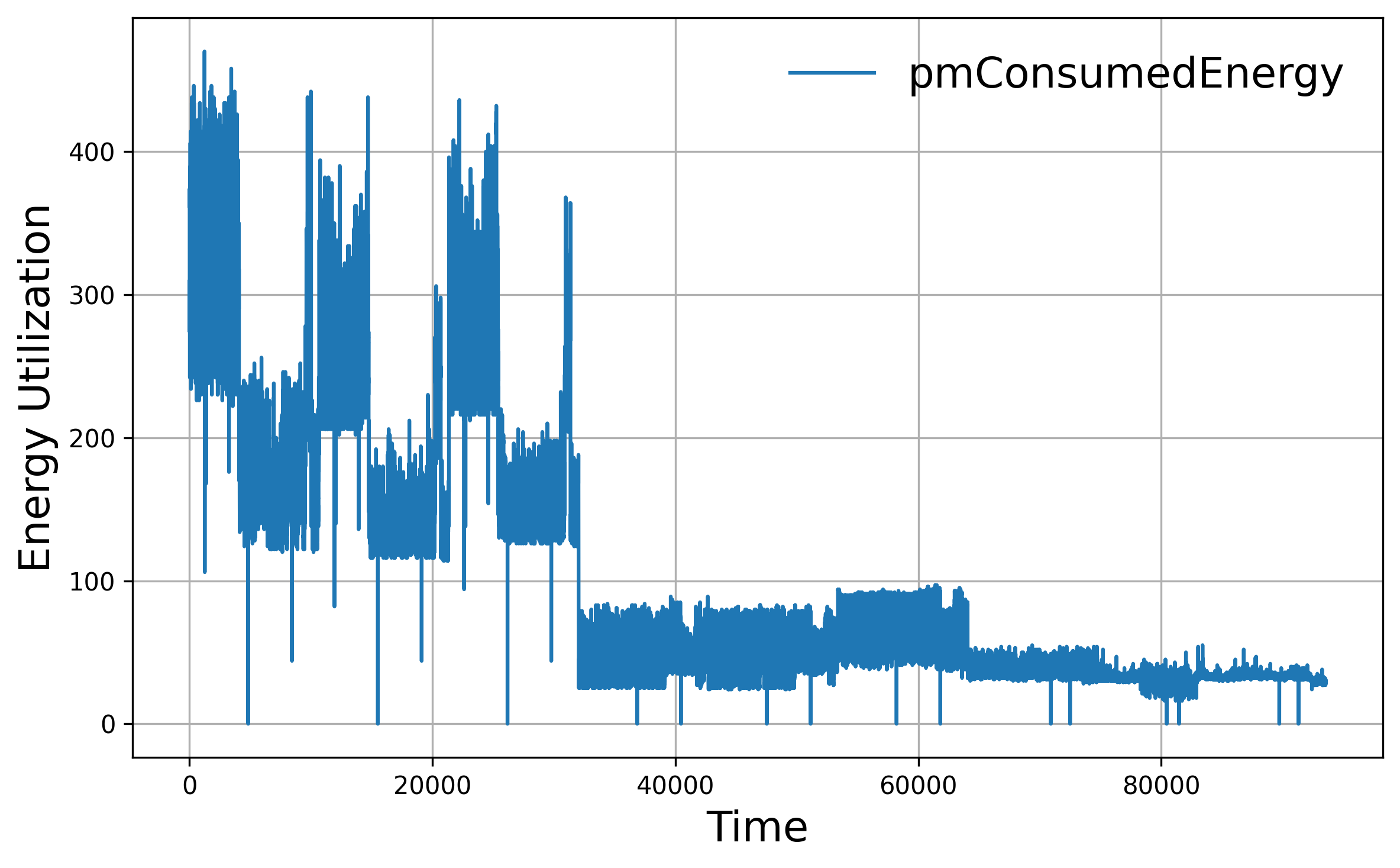}
        \caption{Vendor Data}
        \label{fig:pattern_b}
    \end{subfigure}
    \hfill
    \caption{Snapshot of Internet Activity and Energy Consumption of a Base-station}
    \label{fig:data-pattern}
\end{figure}

\subsection{{Results}}


{The objective of this study is to investigate whether 
ML models can maintain comparable predictive performance when trained on a reduced set of \textit{important samples}, selected using a gradient-based influence metric, compared to training on the full dataset. We have chosen a long-short term memory (LSTM) neural network as a benchmark for the first two datasets to compare with prior art implementation  
\cite{jaffry2020cellular}.
For the third dataset we use the implementation and baseline from \cite{Klautau18}. 
We sort training samples by their importance scores in descending order and select the top $p\%$ for retraining, where $p \in \{10, 20, \ldots, 90\}$. For each selected subset, the LSTM is reinitialized and trained solely on the selected samples and the MAE and training time are recorded. Results are compared to the full model baseline, which is compared to prior published art.}
 
Figure~\ref{fig:test_predictions_all} presents the test prediction plots, comparing the baseline with proposed model predictions for the first two datasets.
In each case, the ground truth (blue), best important samples model (red), and full model (green) are plotted over time, showing how closely the proposed model tracks real-world internet activity. Insets in the plots provide zoomed views of regions with high activity variance, illustrating fine-grained prediction. It turns out that the full model and the one based on the best important samples are almost super-imposed. This framework also captures extreme drift in the Vendor data.

\begin{figure}[htbp]
    \centering
    \begin{subfigure}{0.48\linewidth}
        \includegraphics[width=\linewidth]{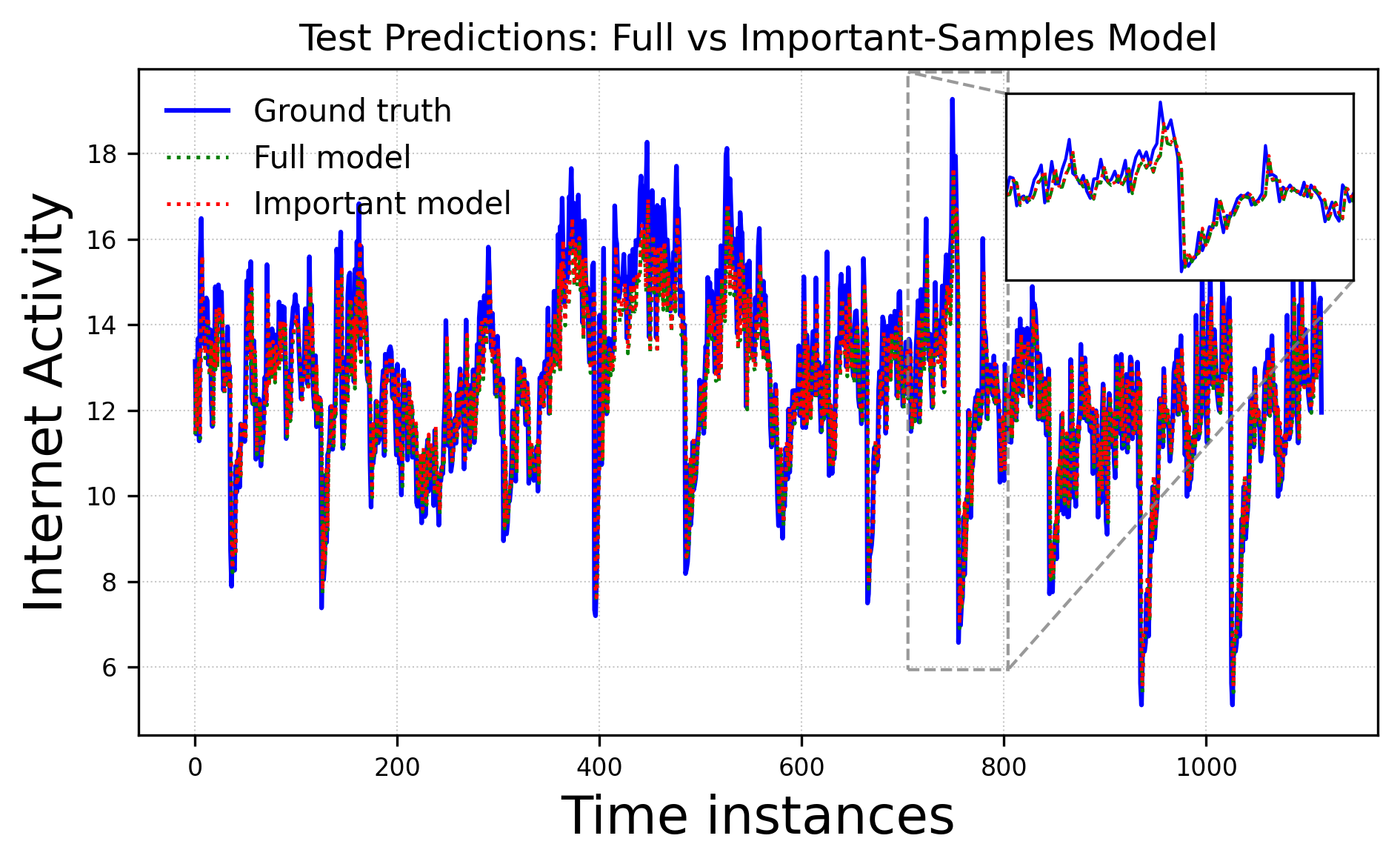}
        \caption{Telecom Italia}
        \label{fig:test_predictions_a}
    \end{subfigure}
    \hfill
    \begin{subfigure}{0.48\linewidth}
        \includegraphics[width=\linewidth]{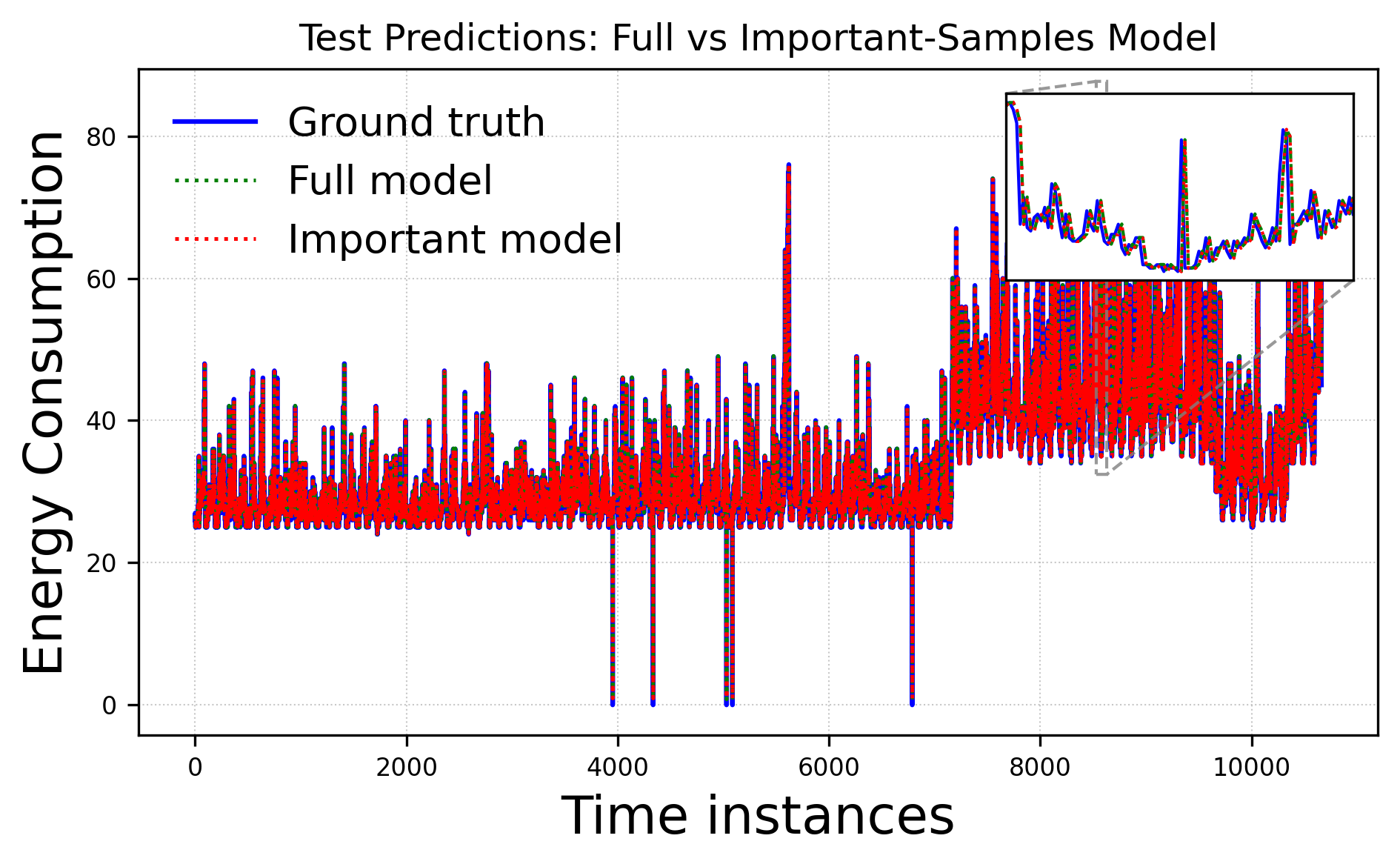}
        \caption{Vendor Data}
        \label{fig:test_predictions_b}
    \end{subfigure}
    \hfill
    
    \caption{Test predictions comparing the baseline model and the sample important model}
    \label{fig:test_predictions_all}
\end{figure}

Figure~\ref{fig:mse_all}a and Figure~\ref{fig:mse_all}b show the Mean Absolute Error (MAE) across five runs as a function of the percentage of top-ranked training samples used. Figure~\ref{fig:mse_all}c and Figure~\ref{fig:mse_all}d show the Root Mean Squared Error (RMSE) difference in azimuth and elevation angles in degrees. The curves demonstrate that the proposed sample selection method achieves accuracy comparable to the full model once a sufficient proportion of informative samples is included, while requiring fewer samples. This leads to notable computational and energy efficiency gains without significant accuracy loss.

\begin{figure}[htbp]
    \begin{subfigure}{0.48\linewidth}
        \includegraphics[width=\linewidth]{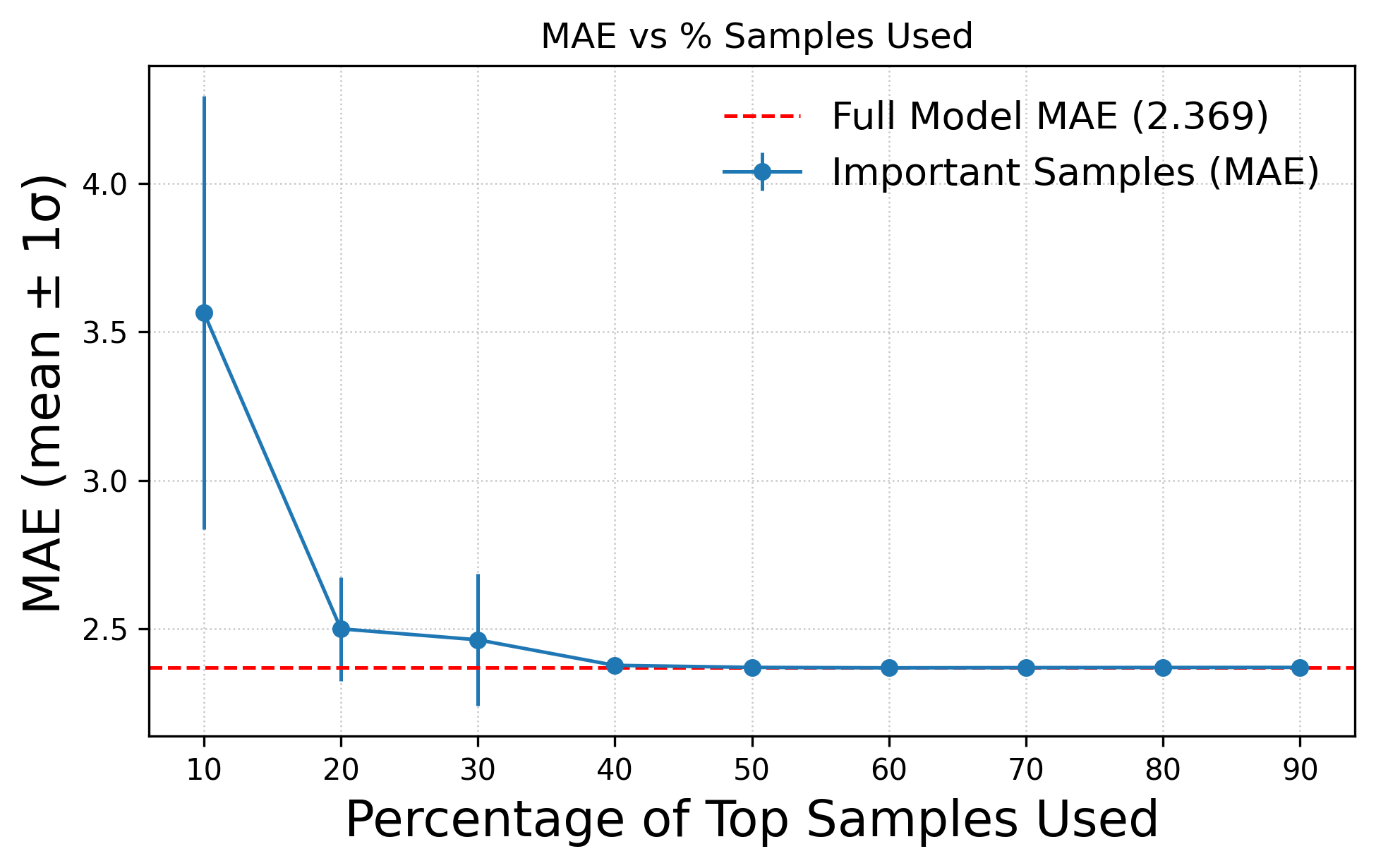}
        \caption{Telecom Italia}
        \label{fig:mse_a}
    \end{subfigure}
    \begin{subfigure}{0.48\linewidth}
        \includegraphics[width=\linewidth]{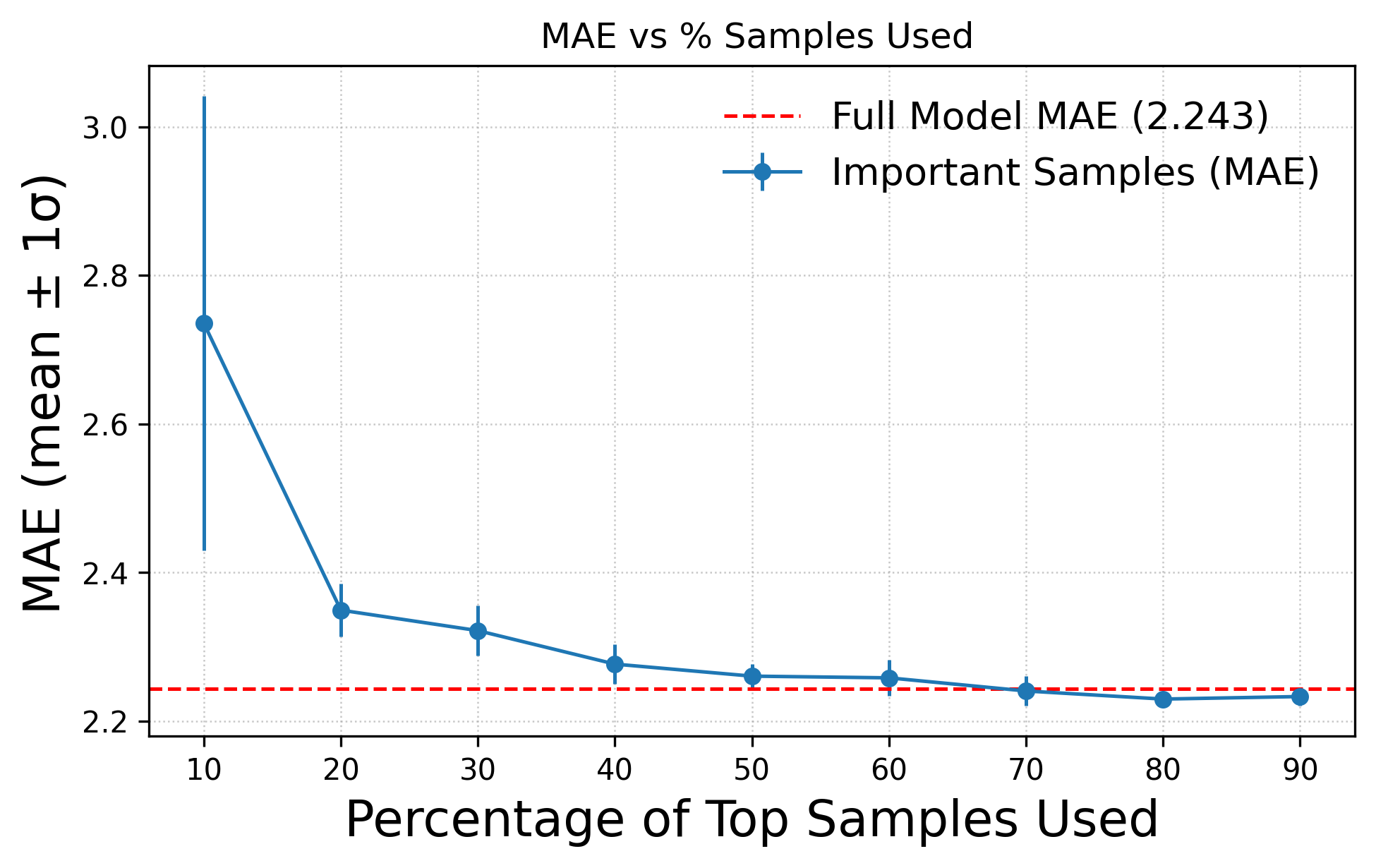}
        \caption{Vendor data}
        \label{fig:mse_c}
    \end{subfigure}    
    \begin{subfigure}{0.48\linewidth}
        \includegraphics[width=\linewidth]{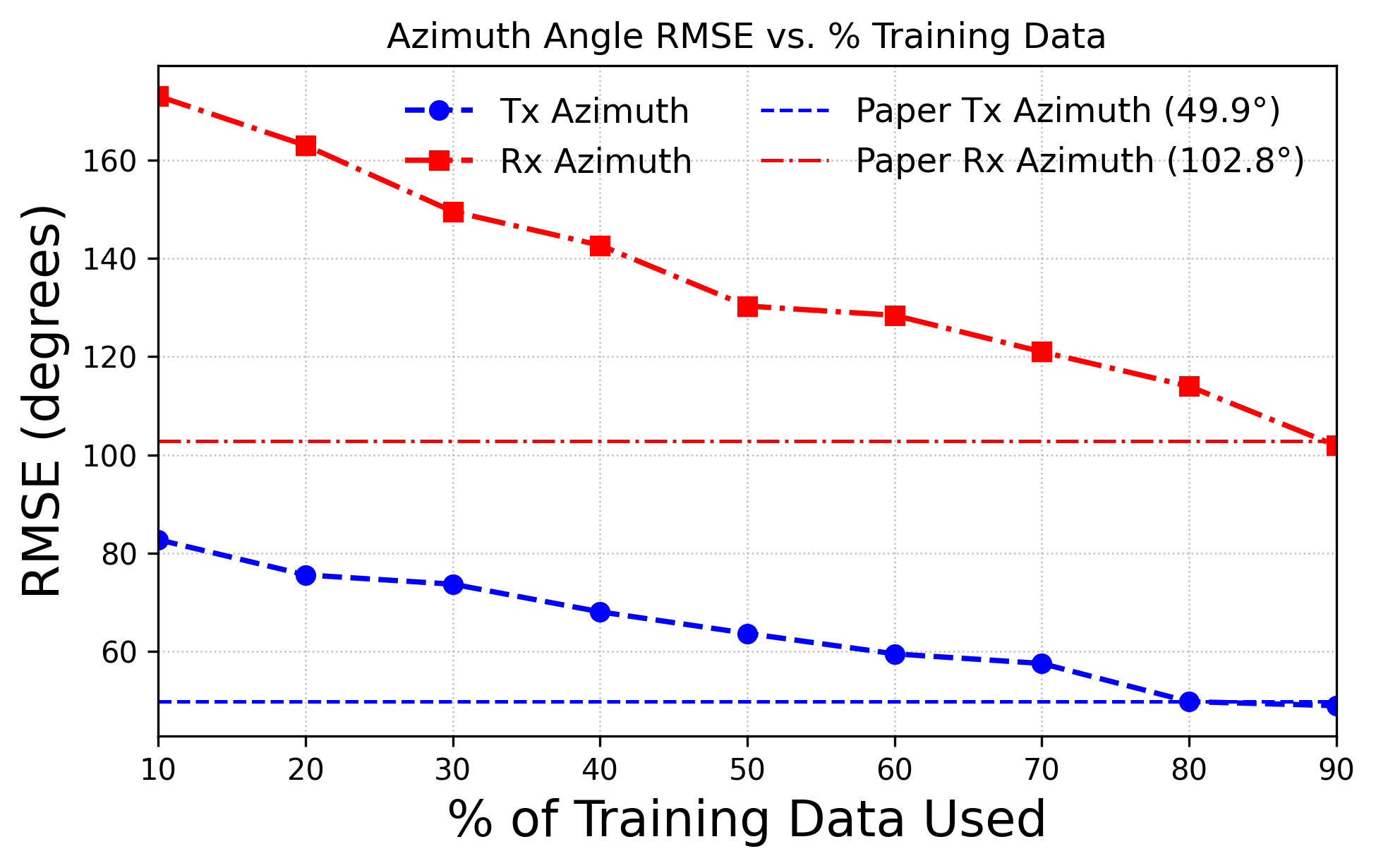}
        \caption{5G Beam Selection(Azimuth Angle)}
        \label{fig:mse_c}
    \end{subfigure}
    \begin{subfigure}{0.48\linewidth}
        \includegraphics[width=\linewidth]{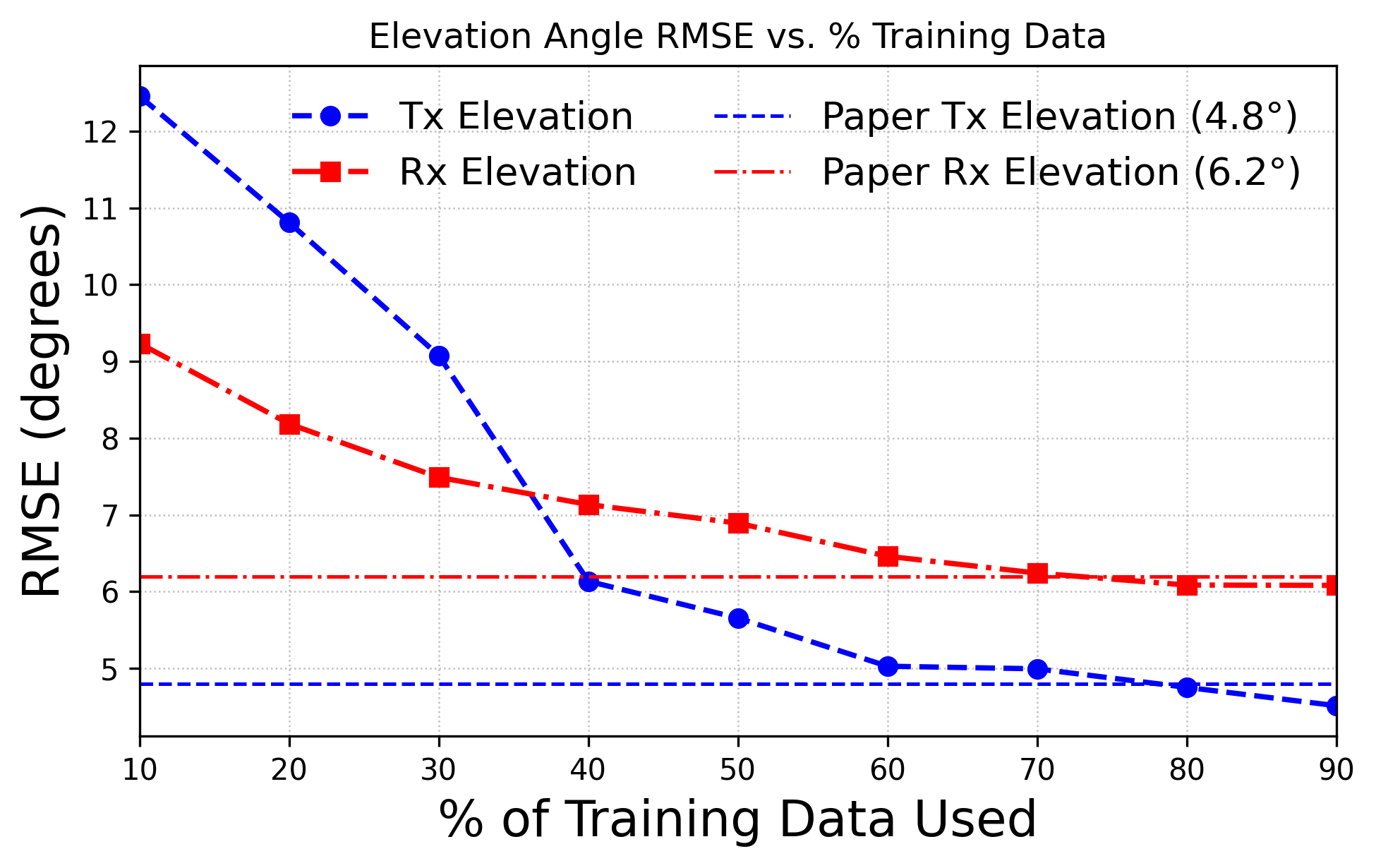}
        \caption{5G Beam Selection (Elevation Angle)}
        \label{fig:mse_c}
    \end{subfigure}
    \caption{Model performance of sample importance framework as compared to baseline models.}
    \label{fig:mse_all}
\end{figure}

For the Telecom Italia dataset, comparable performance is reached with 68\% of important samples, reducing the training data by 28\%. For the Vendor dataset, 74\% suffices to match the baseline, giving a 23\% reduction. In the 5G beam selection dataset, the method outperforms the baseline on elevation angles using 75\% of training samples, and matches the performance  on azimuth angles with 90\% data. Dataset reduction is defined as the gap between the total training data and the subset of important samples selected via gradient-based filtering. Thus, using 78\% of important samples does not imply a 78\% reduction, but rather reflects the proportion of the selected subset used. For statistical significance, results are averaged over $N_{\text{RUNS}} = 5$ independent runs, with error bars denoting mean ± $1\sigma$. A bootstrap 95\% confidence interval (CI) is reported for the improvement in MAE between the important-sample method and the full model, ensuring robustness without assuming strict normality.

Finally, Table~\ref{tab:dataset-improvements} summarizes the improvement in model performance, the percentage of time saved, and the proportion of data required to achieve these results with varying dataset sizes, the latter being selected from the best improvements from Figure~\ref{fig:mse_all}. It turns out that this optimal proportion of data decreases with the model size, and thus so do its improvements in time and accuracy, showing that larger datasets tend to exhibit redundancy.

\begin{table}
  \caption{Model Performance and Training Time Improvements for Different Telecom Italia Dataset Sizes}
  \label{tab:dataset-improvements}
  \centering
  \begin{tabular}{x{2cm}x{2.7cm}x{3cm}x{3.3cm}}
    \toprule
    Dataset Size (Samples) & Percentage of Samples Used (\%) & Model Performance Improvement (MAE) & Training Time  Improvement (seconds) \\
    \midrule
    5K   & 80 & \ \ 2.71  & \ \ 3.07  \\
    50K  & 70 & 21.03 & 15.55 \\
    300K & 65 & 28.35 & 21.03 \\
    \bottomrule
  \end{tabular}
\end{table}

{We monitored the carbon emissions of our experiments using CodeCarbon \cite{benoit_courty_2024_11171501} across the Telecom Italia, Vendor Data, and 5G Beam Selection datasets to assess whether strategically reducing training samples influences energy usage and contributes to broader sustainability goals. Emissions were measured and averaged across multiple runs for each dataset. The important-samples model consistently lowers emissions versus the full model. The Telecom Italia dataset drops from \SI{2.1066e-6}{\kilogram} to \SI{1.3031e-6}{\kilogram} (\SI{-38.14}{\percent}),
Vendor Data from \SI{2.0734e-6}{\kilogram} to \SI{1.2666e-6}{\kilogram} (\SI{-38.91}{\percent}),
and 5G Beam Selection from \SI{1.9918e-6}{\kilogram} to \SI{1.6927e-6}{\kilogram} (\SI{-15.02}{\percent}).
The findings indicate an average reduction of approximately $30.69\%$, demonstrating that sample importance not only enhances computational efficiency but also delivers measurable sustainability benefits in large-scale data analysis workflows.}

\section{Conclusion and future works}

To the best of our knowledge, this is the first study to apply an important-sample–based training strategy in the telecommunication domain. Unlike typical machine learning datasets, telecom data is tabular, time-ordered, and shaped by both seasonal patterns and sudden anomalies. These behaviors appear at fine temporal resolutions, yet operational constraints make it impractical to maintain separate models for different times of day. Hence, adaptive methods that generalize across traffic conditions without full retraining are needed. Our approach addresses this by selectively training on the most impactful data points. This model-agnostic strategy provides a resource-efficient and sustainable way to process large-scale telecom datasets. Evaluations on LSTM-based models show that while small datasets yield limited gains, larger datasets benefit substantially in training efficiency, reduced computation time, and sustained or even improved accuracy. The results confirm that training on carefully chosen samples is more effective than using all data indiscriminately and offer guidance for balancing accuracy against efficiency when slight performance trade-offs are acceptable.

This work represents a conceptual rather than algorithmic contribution, focusing on empirical insight and practical utility over theoretical development. Future research could extend this direction by linking gradient-norm statistics to generalization behavior, incorporating dynamic or curriculum-style selection mechanisms, and benchmarking against more advanced core-set or influence-based approaches under comparable computational budgets. Further validation across diverse model architectures and telecom tasks would also help assess generality. Finally, translating the observed trade-offs into operational guidelines—such as how to select the optimal data fraction under given latency or cost constraints—remains an open area for exploration.

\bibliography{library}


\end{document}